\documentclass[runningheads,a4paper]{llncs}

\usepackage{amssymb}
\usepackage{graphicx}

\usepackage{url}
\urldef{\mailsa}\path|{barta,| \urldef{\mailsb}\path|nagyg,| \urldef{\mailsc}\path|kazi,|
\urldef{\mailsd}\path|henk}@tmit.bme.hu|
\newcommand{\keywords}[1]{\par\addvspace\baselineskip
\noindent\keywordname\enspace\ignorespaces#1}

\begin{document}

\mainmatter  

\title{GEFCOM 2014 - Probabilistic Electricity Price Forecasting}

\titlerunning{GEFCOM 2014 - Probabilistic Electricity Price Forecasting}

\author{Gergo Barta\inst{1} \and Gyula Borbely\and Gabor Nagy\inst{1}\and
Sandor Kazi\inst{1}\and Tamas Henk PhD.\inst{1}}
\authorrunning{GEFCOM 2014 - Probabilistic Electricity Price Forecasting}

\institute{Department of Telecommunications and Media Informatics, Budapest
University of Technology and Economics, Magyar tudosok krt. 2. H-1117 Budapest,
Hungary\\
\mailsa\mailsb\mailsc\mailsd\\
\url{http://www.tmit.bme.hu}}


\toctitle{GEFCOM 2014}
\tocauthor{Probabilistic Electricity Price Forecasting}
\maketitle

\begin{abstract}
Energy price forecasting is a relevant yet hard task in the field of multi-step time series forecasting. In this paper we compare a well-known and established method, ARMA with exogenous variables with a relatively new technique Gradient Boosting Regression. The method was tested on data from Global Energy Forecasting Competition 2014 with a year long rolling window forecast. The results from the experiment reveal that a multi-model approach is significantly better performing in terms of error metrics. Gradient Boosting can deal with seasonality and auto-correlation out-of-the box and achieve lower rate of normalized mean absolute error on real-world data.
\end{abstract}

\section{Introduction}

Forecasting electricity prices is a difficult task as they reflect the actions
of various participants both inside and outside the market. Both producers and
consumers use day-ahead price forecasts to derive their unique strategies and
make informed decisions in their respective businesses and on the electricity
market. High precision short-term price forecasting models are beneficial in
maximizing their profits and conducting cost-efficient business. Day-ahead
market forecasts also help system operators to match the bids of both generating
companies and consumers and to allocate significant energy amounts ahead of
time.

The methodology of the current research paper originates from the GEFCOM 2014
forecasting contest. In last year's contest our team achieved a high ranking
position by ensembling multiple regressors using the Gradient Boosted Regression
Trees paradigm. Promising results encouraged us to further explore potential of
the initial approach and establish a framework to compare results with one of
the most popular forecasting methods; ARMAX.

Global Energy Forecasting Competition is a well-established competition first
announced in 2012 \cite{hong2012} with worldwide success. The 2014 edition
\cite{hong2014} put focus on renewal energy sources and probabilistic
forecasting. The GEFCOM 2014 Probabilistic Electricity Price Forecasting Track
offered a unique approach to forecasting energy price outputs, since competition
participants needed to forecast not a single value but a probability
distribution of the forecasted variables. This methodological difference offers
more information to stakeholders in the industry to incorporate into their daily
work. As a side effect new methods had to be used to produce probabilistic
forecasts.

The report contains five sections: 

\begin{enumerate}
  \item Methods show the underlying models in detail with references.
  \item Data description provides some statistics and description about the target variables and the features used in research.
  \item Experiment Methodology summarizes the training and testing environment and evaluation scheme the research was conducted on.
  \item Results are presented in a the corresponding section.
  \item Conclusions are drawn at the end. 
\end{enumerate}

\section{Methods}
Previous experience showed us that oftentimes multiple regressors are better
than one\cite{rokach}. Therefore we used an ensemble method that was successful
in various other competitions: Gradient Boosted Regression
Trees\cite{aggarwal,mcmahan,graepel}. Experimental results were benchmarked
using ARMAX; a model widely used for time series regression. GBR implementation
was provided by Python's Scikit-learn\cite{skl} library and ARMAX by
Statsmodels\cite{stm}.

\subsection{ARMAX}

We used ARMAX to benchmark our methods because it is a widely applied
methodology for time series regression
\cite{tan2010,feng2010,hou2010,yang2009,hong1996}. This method expands the ARMA
model with (a linear combination of) exogenic inputs (X). ARMA is an
abbreviation of auto-regression (AR) and moving-average (MA). ARMA models were
originally designed to describe stationary stochastic processes in terms of AR
and MA to support hypothesis testing in time series analysis \cite{whittle51}.
As the forecasting task in question has exogenic inputs by specification
therefore ARMAX is a reasonable candidate to be used as a modeler.

Using the ARMAX model (considering a linear model wrt. the exogenous input) the
following relation is assumed and modeled in terms of $X_t$ which is the
variable in question at the time denoted by $t$. According to this the value of
$X_{t}$ is a combination of $\mbox{AR}(p)$ (auto-regression of order $p$),
$\mbox{MA}(q)$ (moving average of order $q$) and a linear combination of the
exogenic input.

\begin{equation} \label{eq:armax}
	X_{t}=\varepsilon_{t}+\sum_{i=1}^{p}\varphi_{i}X_{t-i}+\sum_{i=1}^{q}\theta_{i}\varepsilon_{t-i}+\sum_{i=0}^{b}\eta_{i}d_{t}\left(i\right)
\end{equation}

The symbol $\varepsilon_{t}$ in the formula above represents an error term (generally
regarded as Gaussian noise around zero). $\sum_{i=1}^{p}\varphi_{i}X_{t-i}$
represents the autoregression submodel with the order of $p$: $\varphi_{i}$ is
the $i$-th parameter to weight a previous value. The elements of the sum
$\sum_{i=1}^{q}\theta_{i}\varepsilon_{t-i}$ are the weighted error terms of the
moving average submodel with the order of $q$. The last part of the formula is
the linear combination of exogenic input $d_t$.

Usually $p$ and $q$ are chosen to be as small as they can with an acceptable
error. After choosing the values of $p$ and $q$ the ARMAX model can be trained
using least squares regression to find a suitable parameter setting which
minimizes the error.

\subsection{Gradient Boosting Decision Trees}

Gradient boosting is another ensemble method responsible for combining weak
learners for higher model accuracy, as suggested by Friedman in 2000
\cite{friedman2001}. The predictor generated in gradient boosting is a linear
combination of weak learners, again we use tree models for this purpose.
We iteratively build a sequence of models, and our final predictor will be the
weighted average of these predictors.
Boosting generally results in an additive prediction function:

\begin{equation} \label{eq:boost}
	f^*(X) = \beta_0 + f_1(X_1)+ \ldots + f_p(X_p)
\end{equation}

In each turn of the iteration the ensemble calculates two set of weights:
\begin{enumerate}
  \item one for the current tree in the ensemble
  \item one for each observation in the training dataset
\end{enumerate}

The rows in the training set are iteratively reweighted by upweighting
previously misclassified observations.

The general idea is to compute a sequence of simple trees, where each successive
tree is built for the prediction residuals of the preceding tree.
Each new base-learner is chosen to be maximally correlated with the negative
gradient of the loss function, associated with the whole ensemble.
This way the subsequent stages will work harder on fitting these examples and
the resulting predictor is a linear combination of weak learners.

Utilizing boosting has many beneficial properties; various risk functions are
applicable, intrinsic variable selection is carried out, also resolves
multicollinearity issues, and works well with large number of features without
overfitting.

\section{Data description}
The original competition goal was to predict hourly electricity prices for every
hour on a given day. The provided dataset contained information about the prices
on hourly resolution for a roughly 3 year long period between 2011 and 2013 for
an unknown zone.
Beside the prices two additional variables were in the dataset. One was the
Forecasted Zonal Load ($'z'$) and the other was the Forecasted Total Load
($'t'$). The first attribute is a forecasted electricity load value for the same
zone where the price data came from. The second attribute contains the
forecasted total electricity load in the provider network.
The unit of measurement for these variables remain unknown, as is the precision
of the forecasted values. Also, no additional data sources were allowed to be
used for this competition.

\begin{table} \centering
\caption{Descriptive statistics for the input variables and the target}
\label{tab:inpstat}       
\begin{tabular}{lrrr}
\hline\noalign{\smallskip}
 & Price & Forecasted Total Load & Forecasted Zonal Load \\ 
\noalign{\smallskip}\hline\noalign{\smallskip}
\textbf{count} & 25944 & 25944 & 25944 \\ 
\textbf{mean} & 48.146034 & 18164.103299 & 6105.566181 \\ 
\textbf{std} & 26.142308 & 3454.036495 & 1309.785562 \\ 
\textbf{min} & 12.520000 & 11544 & 3395 \\ 
\textbf{25\%} & 33.467500 & 15618 & 5131 \\ 
\textbf{50\%} & 42.860000 & 18067 & 6075 \\ 
\textbf{75\%} & 54.24 & 19853 & 6713.25 \\ 
\textbf{max} & 363.8 & 33449 & 11441 \\ 
\noalign{\smallskip}\hline
\end{tabular}
\end{table}

\begin{figure} \centering
\includegraphics[height=4.5cm]{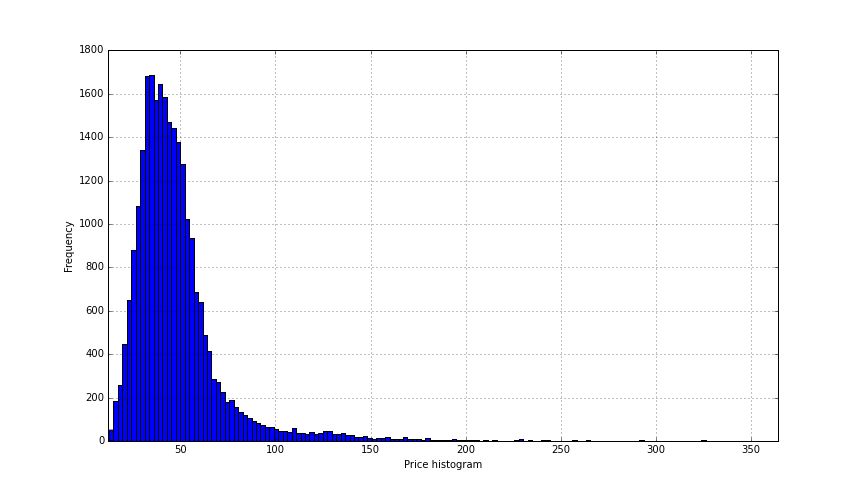}
\caption{Price histogram}
\label{fig:p_hist}
\end{figure}

In Table \ref{tab:inpstat} we can see the descriptive statistic values for the
original variables and the target. The histogram of the target variable (Figure
\ref{fig:p_hist}) is a bit skewed to the left with a long tail on the right and
some unusual high values.
Due to this characteristic we decided to take the natural log value of the
target and build models on that value. The model performance was better indeed
when they were trained on this transformed target.

\begin{figure} \centering
\includegraphics[width=12cm]{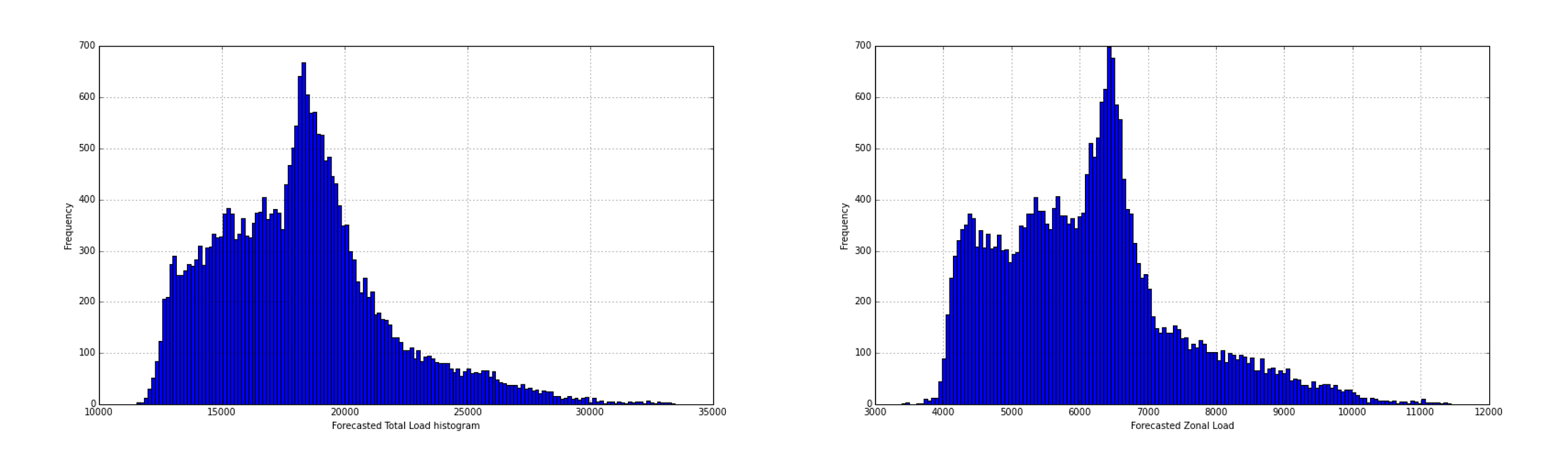}
\caption{Forecasted Total Load and Forecasted Zonal Load histograms}
\label{fig:tz_hist}
\end{figure}

The distribution of the other two descriptive variables are far from normal as we
can see on Figure \ref{fig:tz_hist}. As we can see the shapes are very similar
for these variables with the peak, the left plateau and the tail on the right.
They are also highly correlated with a correlation value of \url{~}0.97, but not
so much with the target itself (\url{~}0.5-0.58).

\begin{table} \centering
\caption{Correlation matrix of input variables}
\label{tab:corrs}       
\begin{tabular}{lrrr}
\hline\noalign{\smallskip}
 & \textbf{Price} & \textbf{Forecasted Zonal Load} & \textbf{Forecasted Total Load} \\ 
\noalign{\smallskip}\hline\noalign{\smallskip}
\textbf{Price} & 1.0 & 0.501915 & 0.582029 \\ 
\textbf{Forecasted Zonal Load} & 0.501915 & 1.0 & 0.972629 \\ 
\textbf{Forecasted Total Load} & 0.582029 & 0.972629 & 1.0 \\ 
\noalign{\smallskip}\hline
\end{tabular}
\end{table}

Beside the variables of Table \ref{tab:inpstat} we also calculated additional
attributes based on them: several variables derived from the two exogenous
variable $'z'$ and $'t'$, also date and time related attributes were extracted
from the timestamps (see Table \ref{tab:attrs} for details).

\bgroup
  \catcode`\_=13%
\begin{table} \centering
\caption{Descriptive features used throughout the competition}
\label{tab:attrs}       
\begin{tabular}{ll}
\hline\noalign{\smallskip}
\textbf{Variable name} & \textbf{Description} \\
\noalign{\smallskip}\hline\noalign{\smallskip}
dow & Day of the week, integer, between 0 and 6 \\
doy & Day of the year, integer, between 0 and 365 \\
day & Day of the month, integer, between 1 and 31 \\
woy & Week of the year, integer, between 1 and 52 \\
hour & Hour of the day, integer, 0-23 \\
month & Month of the year, integer, 1-12 \\
t_M24 & t value from 24 hours earlier \\
t_M48 & t value from 48 hours earlier \\
z_M24 & z value from 24 hours earlier \\
z_M48 & z value from 48 hours earlier \\
tzdif & The difference between t and z \\
tdif & The difference between t and t_M24 \\
zdif & The difference between z and z_M24 \\
\noalign{\smallskip}\hline
\end{tabular}
\end{table}
\egroup

\begin{figure} \centering
\includegraphics[height=5.5cm]{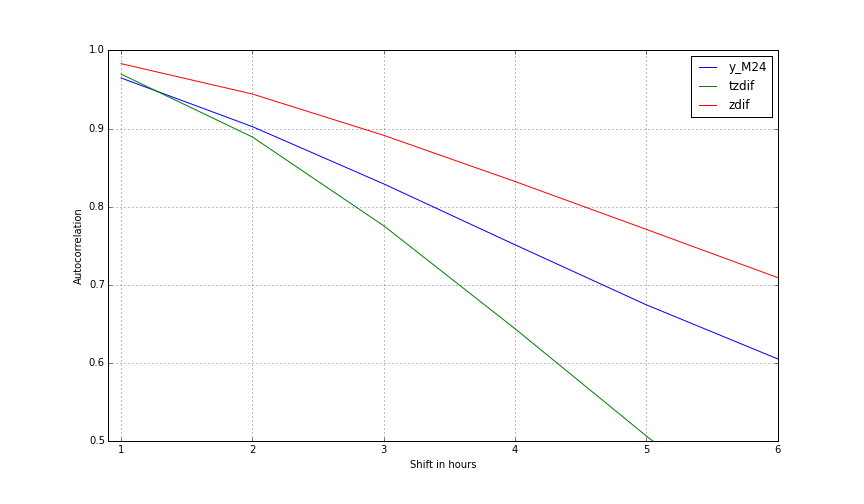}
\bgroup
  \catcode`\_=13%
\caption{Autocorrelation of tzdif, zdif and y_M24 variables}
\egroup
\label{fig:autocorr}
\end{figure}

During the analysis we observed from the
autocorrelation plots that some variables value have stronger correlation with
its +/- 1 hour value, so we also calculated these values for every row. Figure
\ref{fig:autocorr} shows 3 selected variables to be shifted as the
autocorrelation values are extremely high when a lagging window of less than 2
hours is used.

\begin{figure} \centering
\includegraphics[height=6cm]{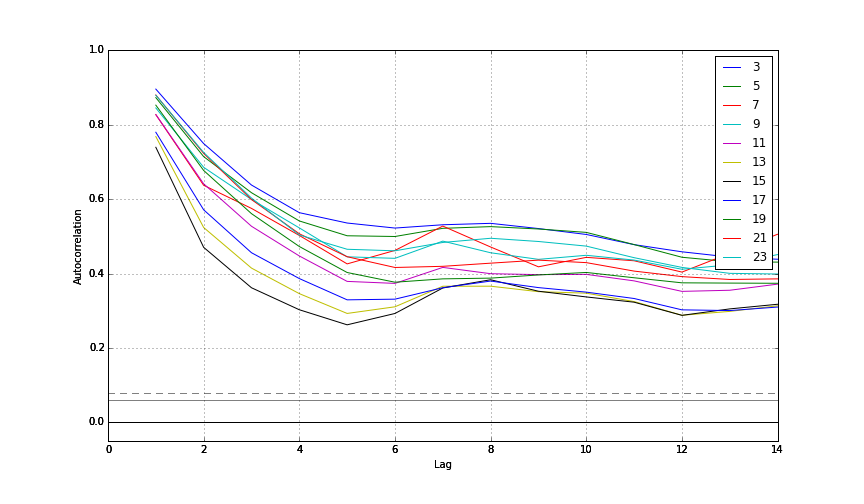}
\caption{Autocorrelation of price values at specific hours, shifted in days}
\label{fig:autocorr2}
\end{figure}

In Figure \ref{fig:autocorr2} figure we can see an autocorrelation plot of price
values in specific hours and they are shifted in days (24 hours). It is clearly
seen that the autocorrelation values for the early and late hours are much
higher than for the afternoon hours. That means it is worth to include shifted
variables in the models as we did. Not surprisingly the errors at the early and
late hours were much lower than midday and afternoon.

Gradient Boosting Regression Trees also provided intrinsic variable importance
measures. Table \ref{tab:attrimportance} shows that (apart from the original
input variables) the calculated differences were found to be important. The
relatively high importance of the hour of day suggests strong within-day
periodicity.

\bgroup
  \catcode`\_=13%
\begin{table} \centering
\caption{Attribute importances provided by GBR}
\label{tab:attrimportance}       
\begin{tabular}{lr}
\hline\noalign{\smallskip}
\textbf{Attribute}  & \textbf{GBR variable importance} \\
\noalign{\smallskip}\hline\noalign{\smallskip}
tzdif & 0.118451 \\
tdif &  0.092485 \\
zdif &  0.090757 \\
z & 0.090276 \\
hour &  0.085597 \\
t & 0.078957 \\
z_M48 &   0.078718 \\
t_M48 &   0.076352 \\
t_M24 &   0.069791 \\
z_M24 &   0.069072 \\
doy &   0.067103 \\
day &   0.056018 \\
dow &   0.024973 \\
month & 0.001449 \\
\noalign{\smallskip}\hline
\end{tabular}
\end{table}
\egroup

\section{Experiment Methodology}
In our research framework we abandoned the idea of probabilistic forecasting as
this is a fairly new approach and our goal was to gain comparable results with
well-established conventional forecasting methods; ARMAX in this case.
 
We used all data from 2013 as a validation set in our research methodology
(unlike in the competition where specific dates were marked for evaluation in
each task). To be on a par with ARMAX we decided to use a rolling window of 30
days to train GBR. This means much less training data (a substantial drawback
for the GBR model), but yields comparable results between the two methods.

The target variable is known until 2013-12-17, leaving us with 350 days for
testing. For each day the training set consisted of the previous 1 month period,
and the subsequent day was used for testing the 24 hourly forecasts. On some
days the ARMAX model did not converge leaving us with 347 days in total to be
used to assess model performance. The forecasts are compared to the known target
variable, we provide 2 metrics to compare the two methods:
Mean Absolute Error (MAE) and Root Mean Squared Error (RMSE). Gradient Boosting
and ARMAX optimizes Mean Squared Error directly meaning that one should focus
more on RMSE than MAE.

\section{Results}
\begin{figure} \centering
\includegraphics[height=6cm]{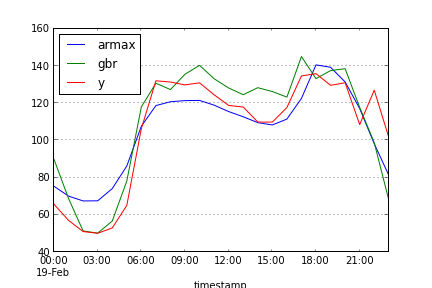}
\bgroup
  \catcode`\_=13%
\caption{Within-day price forecasts for 2013-02-19}
\egroup
\label{fig:rmse_plot}
\end{figure}

Figure \ref{fig:rmse_plot} compares the model outputs with actual prices for a single
day. While Table \ref{tab:errors} shows the descriptive statistics of the error
metrics:
$mae\_p\_armax$, $rmse\_p\_armax$, $mae\_p\_gbr$ and  $rmse\_p\_gbr$ are the
Mean Absolute Errors and Root Mean Squared Errors of ARMAX and GBR models
respectively. The average of the 24 forecasted observations are used for each
day, and the average of daily means are depicted for all the 347 days. In terms
of both RMSE and MAE the average and median error is significantly lower for the
GBR model; surpassing ARMAX by approx. 20\% on average.

During the evaluation we came across several days that had very big error
measures, filtering out these outliers represented by the top and bottom 5\% of
the observed errors we have taken a t-test to confirm that the difference
between the two models is indeed significant ($t=2.3187$, $p=0.0208$ for RMSE).


\begin{table} \centering
\caption{Descriptive statistics for the error metrics}
\label{tab:errors}       
\begin{tabular}{lrrrr}
\hline\noalign{\smallskip}
{} &  mae\_p\_armax &  rmse\_p\_armax &   mae\_p\_gbr &  rmse\_p\_gbr \\
\noalign{\smallskip}\hline\noalign{\smallskip}
count &   347 &    347 &  347 &  347 \\
mean  &     8.640447 &     10.395176 &    7.126920 &    8.496357 \\
std   &    11.809438 &     13.822071 &   10.396122 &   11.627084 \\
min   &     1.223160 &      1.781158 &    1.020160 &    1.302484 \\
5.0\%  &     2.083880 &      2.673257 &    1.439134 &    1.785432 \\
50\%   &     5.152181 &      6.088650 &    3.520733 &    4.144649 \\
95\%   &    27.049138 &     31.339932 &   27.171626 &   31.122828 \\
max   &   101.081747 &    106.317998 &   77.819519 &   83.958518 \\
\noalign{\smallskip}\hline
\end{tabular}
\end{table}


\section{Conclusions and future work}
The GEFCOM competition offered a novel way of forecasting; probabilistic
forecasts offer more information to stakeholders and is an approach worth
investigating in energy price forecasting.
Our efforts in the contest were focused on developing accurate forecasts with
the help of well-established estimators in the literature used in a fairly
different context. This approach was capable of achieving roughly
10\textsuperscript{th} place in the GEFCOM 2014 competition Price Track and
performs surprisingly well when compared to the conventional and widespread
benchmarking method ARMAX overperforming it by roughly 20\%. 

The methodology used in this paper can be easily applied in other domains of
forecasting as well. Applying the framework and observing model performance on a
wider range of datasets yields more robust results and shall be covered in
future work.

During the competition we filtered the GBR training set to better represent the
characteristics of the day to be forecasted, which greatly improved model
performance. Automating this process is also a promising and chief goal of
ongoing research.


\begin{thebibliography}{4}

\bibitem{hong2012} Tao Hong, Pierre Pinson and Shu Fan, "Global Energy
Forecasting Competition 2012", International Journal of Forecasting, vol.30,
no.2., pp 357-363, April - June, 2014

\bibitem{hong2014} Tao Hong, "Energy Forecasting: Past, Present and Future", Foresight:
The International Journal of Applied Forecasting, issue 32, pp. 43-48, Winter
2014.

\bibitem{weron} R. Weron, "Electricity price forecasting: A review of the
state-of-the-art," International Journal of Forecasting, vol. 30, pp. 1030-1081,
2014.

\bibitem{rokach} L. Rokach, "Ensemble-based classifiers," Artificial
Intelligence Review, vol. 33, pp. 1-39, 2010.

\bibitem{aggarwal} S. a. S. L. Aggarwal, "Solar energy prediction using linear
and non-linear regularization models: A study on AMS (American Meteorological
Society) 2013--14 Solar Energy Prediction Contest," Energy, 2014.

\bibitem{mcmahan} H. B. e. a. McMahan, "Ad click prediction: a view from the
trenches.," Proceedings of the 19th ACM SIGKDD international conference on
Knowledge discovery and data mining, pp. 1222-1230, 2013.

\bibitem{graepel} T. e. a. Graepel, "Web-scale bayesian click-through rate
prediction for sponsored search advertising in microsoft's bing search engine.,"
Proceedings of the 27th International Conference on Machine Learning, pp. 13-20,
2010.

\bibitem{skl} Pedregosa, Fabian, et al. "Scikit-learn: Machine learning in
Python." The Journal of Machine Learning Research 12 (2011): 2825-2830.

\bibitem{stm} Seabold, Skipper, and Josef Perktold. "Statsmodels: Econometric
and statistical modeling with python." Proceedings of the 9th Python in Science
Conference. 2010.

\bibitem{tan2010} Ian K. T. Tan, Poo Kuan Hoong, and Chee Yik Keong. 2010.
Towards Forecasting Low Network Traffic for Software Patch Downloads: An ARMA
Model Forecast Using CRONOS. In Proceedings of the 2010 Second International
Conference on Computer and Network Technology (ICCNT '10). IEEE Computer
Society, Washington, DC, USA, 88-92. DOI=10.1109/ICCNT.2010.35
http://dx.doi.org/10.1109/ICCNT.2010.35

\bibitem{feng2010} Gao Feng. 2010. Liaoning Province Economic Increasing
Forecast and Analysis Based on ARMA Model. In Proceedings of the 2010 Third
International Conference on Intelligent Networks and Intelligent Systems (ICINIS
'10). IEEE Computer Society, Washington, DC, USA, 346-348.
DOI=10.1109/ICINIS.2010.107 http://dx.doi.org/10.1109/ICINIS.2010.107

\bibitem{hou2010} Yajun Hou. 2010. Forecast on Consumption Gap Between Cities
and Countries in China Based on ARMA Model. In Proceedings of the 2010 Third
International Conference on Intelligent Networks and Intelligent Systems (ICINIS
'10). IEEE Computer Society, Washington, DC, USA, 342-345.
DOI=10.1109/ICINIS.2010.137 http://dx.doi.org/10.1109/ICINIS.2010.137

\bibitem{yang2009} ShuXia Yang. 2009. The Forecast of Power Demand Cycle Turning
Points Based on ARMA. In Proceedings of the 2009 Second International Workshop
on Knowledge Discovery and Data Mining (WKDD '09). IEEE Computer Society,
Washington, DC, USA, 308-311. DOI=10.1109/WKDD.2009.140
http://dx.doi.org/10.1109/WKDD.2009.140

\bibitem{hong1996} Hong-Tzer, Y. (1996). Identification of ARMAX model for short
term load forecasting: an evolutionary programming approach

\bibitem{whittle51} Whittle, P. (1951). Hypothesis testing in time series
analysis. Uppsala: Almqvist \& Wiksells boktr.

\bibitem{friedman2001} J. H. Friedman, "Greedy function approximation: a
gradient boosting machine," Annals of Statistics, pp. 1189-1232, 2001.

\end{thebibliography}
\end{document}